\documentclass{frontiersSCNS} 

\usepackage{url}

\usepackage{todonotes}

\usepackage{caption}
\usepackage{listings}
\copyrightyear{}
\pubyear{}
\def\journal{Neurosciences}
\def\DOI{}
\def\articleType{}

\def\firstAuthorLast{Alexandre Abraham {et~al}} 

\def\Authors{
    Alexandre Abraham\,$^{1,2,*}$,
    Fabian Pedregosa\,$^{1,2}$,
    Michael Eickenberg\,$^{1,2}$,
    Philippe Gervais\,$^{1,2}$,
    Andreas Muller\,$^{3}$,
    Jean Kossaifi\,$^{4}$,
    Alexandre Gramfort\,$^{1,2,5}$,
    Bertrand Thirion\,$^{1,2}$
    and Ga\"el Varoquaux\,$^{1,2}$}
\def\Address{
    $^{1}$Parietal Team, INRIA Saclay-\^{I}le-de-France, Saclay, France\\
    $^{2}$Neurospin, I\,\textsuperscript{2}BM, DSV, CEA, 91191 Gif-Sur-Yvette, France\\
    $^{3}$Institute of Computer Science VI, University of Bonn, Germany\\
    $^{4}$Department of Computing, Imperial College London, U.K.\\
    $^{5}$Institut Mines-Telecom, Telecom ParisTech, CNRS LTCI, 75014 Paris, France}

\def\corrAuthor{Alexandre Abraham}
\def\corrAddress{Parietal Team, INRIA Saclay-\^{I}le-de-France, Saclay, France}
\def\corrEmail{alexandre.abraham@inria.fr}

\begin{document}
\onecolumn
\firstpage{1}

\title[Machine Learning for Neuroimaging with Scikit-Learn]{Machine Learning for Neuroimaging with Scikit-Learn}
\author[\firstAuthorLast ]{\Authors}
\address{}
\correspondance{}
\editor{}
\topic{Research Topic}

\maketitle
\begin{abstract}

\section{}
Statistical machine learning methods are increasingly used for
neuroimaging data analysis. Their main virtue 
is their ability to model high-dimensional datasets, e.g.\ multivariate
analysis of activation images or resting-state time series.
Supervised learning is typically used in \emph{decoding} or
\emph{encoding} settings to relate
brain images to behavioral or clinical observations, while
unsupervised learning can uncover hidden structures in
sets of images (e.g.\ resting state functional MRI) or find
sub-populations in large cohorts. By considering
different functional neuroimaging applications, we illustrate how scikit-learn,
a Python machine learning library, can be used to perform some key
analysis steps. Scikit-learn contains a very large set of statistical
learning algorithms, both supervised and unsupervised, and its
application to neuroimaging data provides a versatile tool to study the
brain.


\tiny
\section{Keywords:} machine learning, statistical learning, neuroimaging,
scikit-learn, Python
\end{abstract}

\section{Introduction}
Interest in applying statistical machine learning to neuroimaging data analysis
is growing. Neuroscientists use it as a powerful, albeit complex, tool for
statistical inference. The tools are developed by
computer scientists who may lack a deep understanding of the
neuroscience questions. This paper
aims to fill the gap between machine learning and neuroimaging by demonstrating
how a general-purpose machine-learning toolbox, scikit-learn, can
provide state-of-the-art methods for neuroimaging analysis while keeping
the code simple and
understandable by both worlds. Here, we focus on software; for a 
more conceptual introduction to machine learning methods in fMRI analysis,
see \cite{pereira2009} or \cite{mur2009}, while \cite{hastie2001}
provides a good reference on machine learning.
We discuss the use of the scikit-learn toolkit as it is a reference
machine learning tool and has and a variety of algorithms that is matched by few
packages,
but also because it is implemented in Python,
and thus dovetails nicely in the rich neuroimaging Python ecosystem.

This paper explores a few applications of statistical learning to
resolve common neuroimaging needs, detailing the corresponding code,
the choice of the methods, and the underlying assumptions. We discuss not only
prediction scores, but also the interpretability of the results, which
leads us to explore the internal model of various methods. 
Importantly, the GitHub repository of the
paper\footnote{\url{http://www.github.com/AlexandreAbraham/frontiers2013}}
provides complete scripts to generate figures. 
The scope of this paper is not to present a neuroimaging-specific library,
but rather
code patterns related to scikit-learn. However, the nilearn library --\url{http://nilearn.github.io}-- is a software
package under development
that seeks to simplify the use of scikit-learn for neuroimaging. Rather
than relying on an immature and black-box library, we prefer here to
unravel simple and didactic examples of code that enable readers to
build their own analysis strategies.

The paper is organized as
follows. After introducing the \emph{scikit-learn} toolbox, we show 
how to prepare the data to apply
\emph{scikit-learn} routines. Then we describe the application of \emph{supervised
learning} techniques to learn the links between brain images and
stimuli. Finally we demonstrate how \emph{unsupervised learning}
techniques can extract useful structure from the images.


\section{Our tools: scikit-learn and the Python ecosystem}

\subsection{Basic scientific Python tools for the neuroimager}

With its mature scientific stack, Python is a growing contender in the
landscape of neuroimaging data analysis with tools such as Nipy
\citep{millman2007analysis} or Nipype \citep{gorgolewski2011}.
The scientific Python libraries used in this paper are:
\begin{itemize}
    \item{\bf NumPy}. Provides the \verb!ndarray! data type to python,
        an efﬁcient $n$-dimensional data representation for
        array-based numerical computation, similar to that used in Matlab
        \citep{vanderwalt2011}. It handles efficient array persistance
        (input and output) and provides basic operations such as dot
        product. Most scientific Python libraries, including scikit-learn, 
	use NumPy arrays
        as input and output data type.

    \item{\bf SciPy}: higher level mathematical functions that operate on ndarrays for
        a variety of domains including linear algebra, optimization and signal
        processing. SciPy is linked to compiled libraries to ensure high
        performances (BLAS, Arpack and MKL for linear algebra and mathematical
        operations).
        Together, NumPy and SciPy provide a robust scientific environment
        for numerical computing and they are the elementary bricks that we use in all our
        algorithms.

    \item{\bf Matplotlib}, a plotting library tightly integrated into the
        scientific Python stack \citep{hunter2007}. It offers publication-quality figures in
        different formats and is used to generate the figures in
	this paper.

    \item{\bf Nibabel}, to access data in neuroimaging file formats.
	We use it at the beginning of all our scripts.
\end{itemize}

\subsection{Scikit-learn and the machine learning ecosystem}

Scikit-learn \citep{pedregosa2011} is a general purpose machine learning
library written in Python.
It provides efficient implementations of state-of-the-art algorithms,
accessible to non-machine learning experts,
and reusable across scientific disciplines and application fields. It also
takes advantage of Python interactivity and modularity to supply fast and easy
prototyping. There is a variety of other learning packages. For instance,
in Python, PyBrain \citep{schaul2010pybrain} is best at neural networks
and reinforcement learning approaches, but its models are fairly black box,
and do not match our need to interpret the results.
Beyond Python, Weka \citep{hall2009weka} is a rich machine learning framework
written in Java,
however it is more oriented toward data mining. 

Some higher level frameworks provides full pipeline to apply machine
learning techniques to neuroimaging. PyMVPA \citep{hanke2009pymvpa} is a
Python packaging that does data preparation, loading and analysis, as well as
result visualization. It performs multi-variate pattern analysis
and can make use of external tools such as R, scikit-learn or Shogun
\citep{sonnenburg2010}.
PRoNTo \citep{schrouff2013pronto} is written in Matlab and can easily
interface with SPM but does not propose many machine learning algorithms.
Here, rather than full-blown neuroimaging analysis pipelines, we discuss
lower-level patterns that break down how neuroimaging
data is input to scikit-learn and processed with it. Indeed, the breadth
of machine learning techniques in scikit-learn and the variety of
possible applications are too wide to be fully exposed in a high-level
interface. Note that a package like PyMVPA that can rely on scikit-learn for
neuroimaging data analysis implements similar patterns behind its
high-level interface.

\subsection{Scikit-learn concepts}
\label{scikitlearn}

In {\em scikit-learn}, all objects and algorithms accept input data in the form of
2-dimensional arrays of size samples $\times$ features.
This convention makes it generic and domain-independent.
Scikit-learn objects share a uniform set of methods that
depends on their purpose: \textit{estimators} can fit models from data,
\textit{predictors} can make predictions on new data and \textit{transformers}
convert data from one representation to another.

\begin{itemize}
\item {\bf Estimator}. The \textit{estimator} interface, the core of the
    library, exposes a \texttt{fit} method for learning model parameters from training data.
    All supervised
    and unsupervised learning algorithms (e.g., for classification, regression or
    clustering) are available as objects implementing this interface. Machine
    learning tasks such as feature selection or dimensionality
    reduction are also provided as estimators.

\item {\bf Predictor}. A \textit{predictor} is an estimator with
    a \texttt{predict}
    method that takes an input array \texttt{X\_test} and makes
    predictions for each sample in it.
    We denote this input parameter ``\texttt{X\_test}'' in order
    to emphasize that \texttt{predict} generalizes to new data. In the case of
    supervised learning estimators, this method typically returns the predicted
    labels or values computed from the estimated model.

\item {\bf Transformer}. As it is common to modify or filter data before feeding it to a learning
    algorithm, some estimators, named \textit{transformers}, implement a
    \texttt{transform} method. Preprocessing, feature selection and
    dimensionality reduction
    algorithms are all provided as transformers within the library. If the transformation
    can be inverted, a method called \texttt{inverse\_transform} also exists.

\end{itemize}

\smallskip

When testing an estimator or setting hyperparameters, one needs a reliable
metric to evaluate its performance. Using the same
data for training and testing is not acceptable because it leads to
overly confident model performance, a phenomenon also known as \emph{overfitting}.
Cross-validation is a technique that allows one to reliably evaluate an
estimator on a given dataset. It consists in iteratively fitting the
estimator on a fraction of the data, called \emph{training set}, and testing it
on the left-out unseen data, called \emph{test set}.
Several strategies exists to partition the data.
For example, $k$-fold cross-validation consists in dividing (randomly or not) the samples in $k$
subsets: each subset is then used once as testing set while the others $k - 1$
subsets are used to train the estimator. This is one of the simplest and most
widely used cross-validation strategies. The parameter $k$ is commonly set
to 5 or 10. Another strategy, sometimes called Monte-Carlo cross-validation,
uses many random partitions in the data.

For a given model and some fixed value of hyperparameters, the scores
on the various test sets can be averaged to give a quantitative score
to assess how good the model is. Maximizing this cross-validation score offers
a principled way to set hyperparameters and allows to choose between
different models. This procedure is known as \emph{model selection}.
In {\em scikit-learn}, hyperparameters tuning can be conviently done with the
\texttt{GridSearchCV} estimator. It takes as input an estimator and
a set of candidate hyperparameters. Cross-validation scores are then
computed for all hyperparameters combinations, possibly in parallel,
in order to find the best one. In this paper, we set the regularization
coefficient with grid search in section~\ref{kamitani}.

\section{Data preparation: from MR volumes to a data matrix}
\label{data_preparation}
Before applying statistical learning to neuroimaging data, standard
preprocessing must be applied. For fMRI, this includes motion
correction, slice timing correction, coregistration with an anatomical image and normalization to a common
template like the MNI (Montreal Neurologic Institute) one if necessary.
Reference softwares for these tasks are
SPM~\citep{friston2007} and
FSL~\citep{smith2004}. A Python
interface to these tools is available in nipype Python library
\citep{gorgolewski2011}. Below we discuss shaping preprocessed data into
a format that can be fed to scikit-learn. For the machine learning
settings, we need a data matrix, that we will denote $X$, and optionally a
target variable to predict, $y$.

\subsection{Spatial resampling}
\label{resampling}

Neuroimaging data often come as Nifti files, 4-dimensional data (3D scans
with time series at each location or voxel) along with a
transformation matrix (called affine) used to compute voxel locations
from array indices to world coordinates. When working with several subjects,
each individual data is registered on a common template (MNI, Talairach...),
hence on a common affine, during preprocessing.

Affine matrix can express data
anisotropy, when the distance between two voxels is not the same
depending on the direction. This information is used by algorithms
relying on the spatial structure of the data, for instance the
Searchlight.

SciPy routine \texttt{scipy.ndimage.affine\_transform}
can be used to perform image resampling: 
changing the spatial resolution of the data\footnote{An easy-to-use
implementation is proposed in nilearn}. This is
an interpolation and alters the data, that is why it should be used carefully.
Downsampling is commonly used to reduce the size of data to process.
Typical sizes are 2mm or 3mm resolution, but scan spatial resolution is
increasing with progress in MR physics. The affine matrix can encode the
scaling factors for each direction.

\subsection{Signal cleaning}

Due to its complex and indirect acquisition process, neuroimaging data often
have a low
signal-to-noise ratio. They contain trends and artifacts that must be removed
to ensure maximum machine learning algorithms efficiency. Signal cleaning
includes:
\begin{itemize}
    \item{\bf Detrending} removes a linear trend over the time series of each
        voxel. This is a useful step when studying fMRI data, as the voxel
        intensity itself has no meaning and we want to study its variation and
        correlation with other voxels. Detrending can be done thanks to SciPy
        (\texttt{scipy.signal.detrend}).
    \item{\bf Normalization} consists in setting the timeseries variance to 1.
        This harmonization is necessary as some machine learning algorithms are
        sensible to different value ranges.
    \item{\bf Frequency filtering} consists in removing high or low
        frequency signals. Low-frequency signals in fMRI data are caused by
        physiological mechanisms or scanner drifts. Filtering can be done thanks
        to a Fourier transform (\texttt{scipy.fftpack.fft}) or a Butterworth
        filter (\texttt{scipy.signal.butter}).
\end{itemize}

\subsection{From 4-dimensional images to 2-dimensional array: masking}

\label{sec:unmasking}

Neuroimaging data are represented in 4 dimensions: 3 spatial dimensions, and 
one dimension to index time or trials.
Scikit-learn algorithms, on the other hand, only accept 2-dimensional
samples $\times$ features matrices (see Section~\ref{scikitlearn}).
Depending on the setting, voxels and time series can be
considered as features
or samples. For example, in spatial independent component analysis (ICA),
voxels are samples.

The reduction process from 4D-images to feature vectors comes with the loss
of spatial structure. It however allows to discard uninformative
voxels, such as the ones outside of the brain. Such voxels that
only carry noise and scanner artifacts would reduce SNR and affect the
quality of the estimation. The selected voxels form a \emph{brain mask}.
Such a mask is often given along with the datasets or can be computed
with software tools such as FSL or SPM.

\begin{figure}[hbtp]
    \begin{center}
        \includegraphics[width=.5\linewidth]{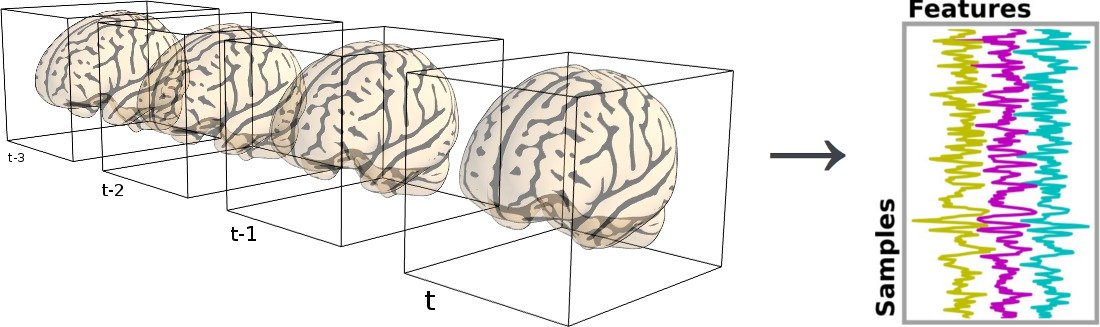}
    \end{center}
    \caption{Conversion of brain scans into 2-dimensional data}
    \label{fig:niimg}
\end{figure}

Applying the mask is made easy by NumPy advanced indexing using boolean arrays.
Two-dimensional masked data will be referred to as \texttt{X} to follow
scikit-learn conventions:
\begin{lstlisting}
mask = nibabel.load('mask.nii').get_data()
func_data = nibabel.load('epi.nii').get_data()
# Ensure that the mask is boolean
mask = mask.astype(bool)
# Apply the mask, X = timeseries * voxels
X = func_data[mask].T

# Unmask data
unmasked_data = numpy.zeros(mask.shape, dtype=X.dtype)
unmasked_data[mask] = X
\end{lstlisting}

\subsection{Data visualisation}

Across all our examples, voxels of interest are represented on an axial slice of
the brain. Some transformations of the original matrix data are required to match
matplotlib data format. The following snippet of code shows how to load and
display an axial slice overlaid with an activation map. The background is an
anatomical scan and its highest voxels are used as synthetic
activations.

\begin{lstlisting}
# Load image
bg_img = nibabel.load('bg.nii.gz')
bg = bg_img.get_data()
# Keep values over 6000 as artificial activation map
act = bg.copy()
act[act < 6000] = 0.

# Display the background
plt.imshow(bg[..., 10].T, origin='lower', interpolation='nearest', cmap='gray')
# Mask background values of activation map
masked_act = np.ma.masked_equal(act, 0.)
plt.imshow(masked_act[..., 10].T, origin='lower', interpolation='nearest', cmap='hot')
# Cosmetics: disable axis
plt.axis('off')
plt.show()
\end{lstlisting}

Note that a background is needed to display partial maps. Overlaying two images
can be done thanks to the \texttt{numpy.ma.masked\_array} data structure.
Several options exist to enhance the overall aspect of the plot.
Some of them can be found in the full scripts provided with this paper.
It generally
boils down to a good knowledge of Matplotlib. Note that the Nipy package provides a
\texttt{plot\_map} function that is tuned to display activation maps (a
background is even provided if needed).

\section{Decoding the mental representation of objects in the brain}

In the context of neuroimaging, \textit{decoding} refers to learning a model
that predicts behavioral or phenotypic variables from brain imaging data. 
The alternative that consists in predicting the imaging data given external
variables, such as stimuli descriptors, is 
called \textit{encoding}~\citep{naselaris2011}. It is further discussed in the next 
section.

First, we illustrate decoding with a simplified version of the experiment presented in
\cite{haxby2001}. In the original work, visual stimuli from 8 different categories
are presented to 6 subjects during 12 sessions. The goal is to 
predict the category of the stimulus presented to the subject given the
recorded fMRI volumes. This example has already been widely analyzed
\citep{hanson2004combinatorial,detre2006multi,otoole2007,hanson2008brain,hanke2009pymvpa} and has become a reference
example in matter of decoding. For the sake of simplicity, we restrict the example
to one subject and to two categories, faces and houses.

As there is a \emph{target} variable $y$ to predict, this is a supervised
learning problem. Here $y$ represents the two object categories, a.k.a.
\emph{classes} in machine-learning terms. In such settings, where $y$
takes discrete values the learning
problem is known as \emph{classification}, as opposed to
\emph{regression} when the variable $y$ can take continuous values,
such as age.

\subsection{Classification with feature selection and linear SVM}

Many classification methods are available in scikit-learn. In this
example we chose to
combine the use of univariate feature selection and Support Vector
Machines (SVM). Such a classification strategy is simple
yet efficient when used on neuroimaging data.

\smallskip

After applying a brain mask, the data consist of 40\,000 voxels, here
the features, for only 1\,400 volumes, here the samples.
Machine learning with many more features than samples
is challenging, due to the so-called \emph{curse of dimensionality}.
Several strategies exist to reduce the number of features.
A first one is based on prior neuroscientific
knowledge. Here one could restrict the mask to occipital areas, where the visual
cortex is located. Feature selection is a second, data-driven, approach
that relies on a univariate
statistical test for each individual feature. Variables
with high individual discriminative power are kept.

Scikit-learn offers a panel of strategies to select features. In supervised
learning, the most popular feature selection method is the
F-test. 
The null hypothesis of this test is that the feature takes the same value
independently of the value of $y$ to predict.
In scikit-learn, \verb!sklearn.feature_selection! proposes a panel
of feature selection strategies. One can choose to take a percentile of the features
(\verb!SelectPercentile!), or a fixed number of features (\verb!SelectKBest!).
All these objects are implemented as transformers (see
 section~\ref{scikitlearn}).
The code below uses the \verb!f_classif! function (ANOVA F-Test) along with
the selection of a fixed number of features.

\smallskip

On the reduced feature set, we use a linear SVM classifier, 
\verb!sklearn.svm.SVC!, to find the hyperplane that maximally
separates the samples belonging to the different classes.
Classifying a new sample boils
down to determining on which side of the hyperplane it lies. With a
linear kernel, the separating hyperplane is defined in the input
data space and its coefficients can be related to the voxels.
Such coefficients can therefore be visualized as an image (after
unmasking step described in~\ref{sec:unmasking})
where voxels with high values have more influence on the prediction
than the others (see figure~\ref{fig:haxby}).

\begin{lstlisting}
feature_selection = SelectKBest(f_classif, k=500)
clf = SVC(kernel='linear')
X_reduced = feature_selection.fit_transform(X)
clf.fit(X_reduced, y)
### Look at the discriminating weights
coef = clf.coef_
### Reverse feature selection
coef = feature_selection.inverse_transform(coef)
\end{lstlisting}

\subsection{Searchlight}
\label{searchlight}

Searchlight \citep{kriegeskorte2006} is a popular algorithm in the
neuroimaging community. It runs a predictive model on a spatial
neighborhood of each voxel and tests the out-of-sample prediction
performance as proxy measure of the link between the local brain activity
and the target behavioral variable. In practice, it entails performing
cross-validation of the model, most often an SVM, on voxels contained in
balls centered on each voxel of interest. The procedure implies 
solving a large number of SVMs and is computationally expensive.
Detailing an efficient implementation of this algorithm is beyond the
scope of this paper. However, code for searchlight and to generate
figure~\ref{fig:haxby}
is available in the GitHub repository accompanying the paper.

\subsection{Results}

\begin{figure}[hbtp]
  \begin{center}
  \includegraphics[width=\linewidth]{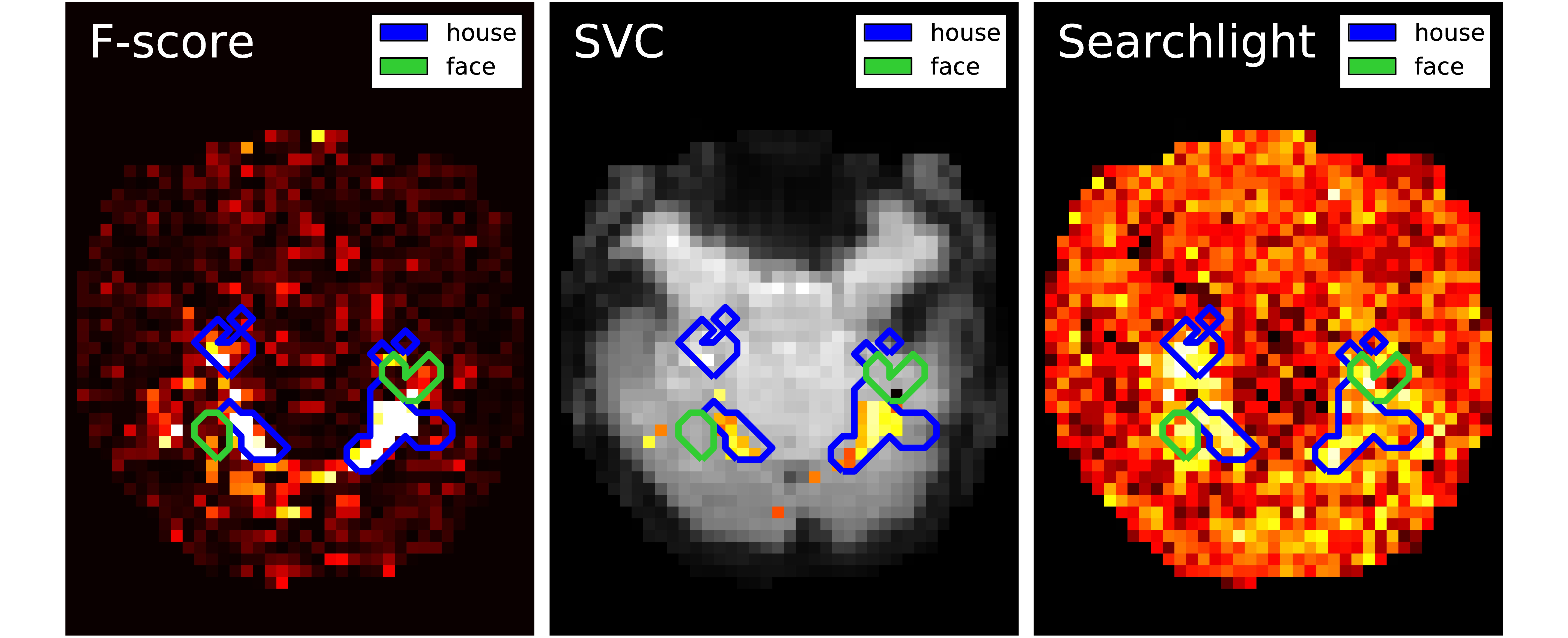}
  \end{center}
\caption{Maps derived by different methods for face versus house
recognition in the Haxby experiment -- \emph{left}: standard analysis;
\emph{center}: SVM weights after screening voxels with an ANOVA; \emph{right}:
Searchlight map. The masks derived from standard analysis in the original
paper \citep{haxby2001} are displayed in blue and green.}
\label{fig:haxby}
\end{figure}

Results are shown in figure~\ref{fig:haxby}: first
F-score, that is standard analysis in brain mapping but also the
statistic used to select features; second the SVC weights after feature
selection and last the Searchlight map.
Note that the voxels with larger weights roughly match for all methods and
are located in the house-responsive areas as defined by the original paper.
The Searchlight is more expanded and blurry than the other methods
as it iterates over a ball around the voxels.

These results match neuroscientific knowledge as they highlight the
high level regions of the ventral visual cortex which is known to
contain category-specific visual areas. While Searchlight only gives a
score to each voxel, the SVC can be used afterward to classify unseen
brain scans.

Most of the final example script (\texttt{haxby\_decoding.py} on GitHub) is
for data
loading and result visualization. Only 5 lines are needed to run a scikit-learn
classifier. In addition, thanks to the scikit-learn modularity, the SVC can be easily
replaced by any other classifier in this example. As all linear models share the
same interface, replacing the SVC by another linear model, such as ElasticNet
or LogisticRegression, requires changing only one line. Gaussian Naive Bayes is a non-linear
classifier that should perform well in this case, and modifiying display can be
done by replacing \texttt{coef\_} by \texttt{theta\_}.

\section{Encoding brain activity and decoding images}
\label{kamitani}

In the previous experiment, the category of a visual stimulus was inferred from
brain activity measured in the visual cortex.
One can go further by inferring a direct link between the image
seen by the subject and the associated fMRI data.

In the experiment of \cite{miyawaki2008} several series of $10{\times}10$
binary images are presented to two subjects while activity on the visual cortex
is recorded.
In the original paper, the training set is composed of random images (where black and white pixels
are balanced) while the testing set is composed of structured images containing
geometric shapes (square, cross...) and letters. Here, for the sake of simplicity, we consider only the training set and use cross-validation to
obtain scores on unseen data.
In the following example, we study the relation between stimuli pixels and
brain voxels in both directions: the reconstruction of the visual stimuli
from fMRI, which is a decoding task, and the prediction of fMRI data
from descriptors of the visual stimuli, which is an encoding task.

\subsection{Decoding}

In this setting, we want to infer the binary visual stimulus presented to
the subject from the recorded fMRI data.
As the stimuli are binary, we will treat this problem as a classification
problem. This implies that the method presented here cannot be extended as-is to
natural stimuli described with gray values. 

In the original work, \cite{miyawaki2008} uses a Bayesian logistic regression
promoting sparsity along with a sophisticated multi-scale strategy.
As one can indeed expect the number of predictive voxels to be limited, we 
compare the $\ell_2$ SVM used above with
a logistic regression and a SVM
penalized with the $\ell_1$ norm
known to promote sparsity. The $\ell_1$ penalized SVM classifier compared here
uses a square-hinge loss while the logistic regression uses a logit function.


\begin{table}[htbp]
    \begin{center}
    \begin{tabular}{l|llllll}
        $C$ value           & 0.0005 & 0.001  & 0.005  & 0.01   & 0.05    & 0.1    \\
    \hline\\[-.9em]
    \hline\\[-.7em]
        $\ell_1$ Logistic Regression & 0.50 $\pm$ .02
                                     & 0.50 $\pm$ .02 
                                     & 0.57 $\pm$ .13
                                     & 0.63 $\pm$ .11
                                     & \textbf{0.70} $\pm$ .12
                                     & 0.70 $\pm$ .12 \\[.1em]
        $\ell_2$ Logistic Regression & 0.60 $\pm$ .11
                                     & 0.61 $\pm$ .12 
                                     & 0.63 $\pm$ .13
                                     & 0.63 $\pm$ .13
                                     & \textbf{0.64} $\pm$ .13
                                     & 0.64 $\pm$ .13 \\[.1em]
        $\ell_1$ SVM classifier (SVC)& 0.50 $\pm$ .06
                                     & 0.55 $\pm$ .12
                                     & 0.69 $\pm$ .11
                                     & \textbf{0.71} $\pm$ .12
                                     & 0.69 $\pm$ .12
                                     & 0.68 $\pm$ .12 \\[.1em]
        $\ell_2$ SVM classifier (SVC)& 0.67 $\pm$ .12
                                     & \textbf{0.67} $\pm$ .12
                                     & 0.67 $\pm$ .12
                                     & 0.66 $\pm$ .12
                                     & 0.65 $\pm$ .12
                                     & 0.65 $\pm$ .12
    \end{tabular}
    \end{center}
    \caption{5-fold cross validation accuracy scores obtained for different
    values of parameter $C$ ($\pm$ standard deviation).}
    \label{fig:miyawaki_cv}
\end{table}

Table~\ref{fig:miyawaki_cv} reports the performance of the different classifiers
for various values of C using a 5-fold cross-validation.
We first observe that setting the parameter $C$ is
crucial as performance drops for inappropriate values of C. It is particularly
true for $\ell_1$ regularized models.
Both $\ell_1$ logistic regression and SVM yield similar performances,
which is not surprising as they implement similar models.

\begin{lstlisting}
from sklearn.linear_model import LogisticRegression as LR
from sklearn.cross_validation import cross_val_score

pipeline_LR = Pipeline([('selection', SelectKBest(f_classif, 500)),
                        ('clf', LR(penalty='l1', C=0.05)])

scores_lr = []
# y_train = n_samples x n_voxels
# To iterate on voxels, we transpose it.
for pixel in y_train.T:
    score = cross_val_score(pipeline_LR, X_train, pixel, cv=5)
    scores_lr.append(score)
\end{lstlisting}

\subsection{Encoding}
Given an appropriate model of the stimulus, e.g. one which can provide an
approximately linear representation of BOLD activation, an encoding approach
allows one to quantify for each voxel to what extent its variability is captured
by the model. A popular evaluation method is the predictive \(r^2\) score, which
uses a prediction on left out data to quantify the decrease in residual norm 
brought about by fitting a regression function as opposed to fitting a constant. 
The remaining variance consists of potentially unmodelled, but reproducible signal
and spurious noise.

On the Miyawaki dataset, we can observe that mere black and white pixel values
can explain a large part of the BOLD variance in many visual voxels. Sticking
to the notation that \(X\) represesents BOLD signal and \(y\) the stimulus, we
can write an encoding model using the ridge regression estimator:

\begin{lstlisting}
from sklearn.linear_model import Ridge
from sklearn.cross_validation import KFold

cv = KFold(len(y_train), 10)
# Fit ridge model, calculate predictions on left out data
# and evaluate r^2 score for each voxel
scores = []
for train, test in cv:
    pred = (Ridge(alpha=100.).fit(y_train[train], X_train[train])
                             .predict(y_train[test]))
    X_true = X_train[test]
    scores.append(
        1. - ((X_true - pred) ** 2).sum(axis=0) /
        ((X_true - X_true.mean(axis=0)) ** 2).sum(axis=0))

mean_scores = np.mean(scores, axis=0)
\end{lstlisting}

Note here that the Ridge can be replaced by a Lasso estimator, which can
give better prediction performance at the cost of computation time.

\subsubsection{Receptive fields}
Given the retinotopic structure of early visual areas, it is expected
that the voxels well predicted by the presence of a black or white pixel
are strongly localized in so-called population
receptive fields (\textit{prf}). This suggests that only very few
stimulus pixels should suffice to explain the activity in each brain
voxel of the posterior visual cortex.
This information can be exploited by using a sparse linear
regression --the Lasso~\citep{tibshirani:96}-- to find the receptive fields.
Here we use the \emph{LassoLarsCV} estimator that relies on the LARS
algorithm~\citep{Efron04leastangle} and
cross-validation to set the Lasso parameter.

\begin{lstlisting}
from sklearn.linear_model import LassoLarsCV

# choose number of voxels to treat, set to None for all voxels
n_voxels = 50
# choose best voxels
indices = mean_scores.argsort()[::-1][:n_voxels]

lasso = LassoLarsCV(max_iter=10)

receptive_fields = []
for index in indices:
    lasso.fit(y_train, X_train[:, index])
    receptive_fields.append(lasso.coef_.reshape(10, 10))

\end{lstlisting}

\subsection{Results}

\label{sec:miyawaki_results}
\begin{figure}[hbtp]
  \begin{center}
    \includegraphics[width=\linewidth]{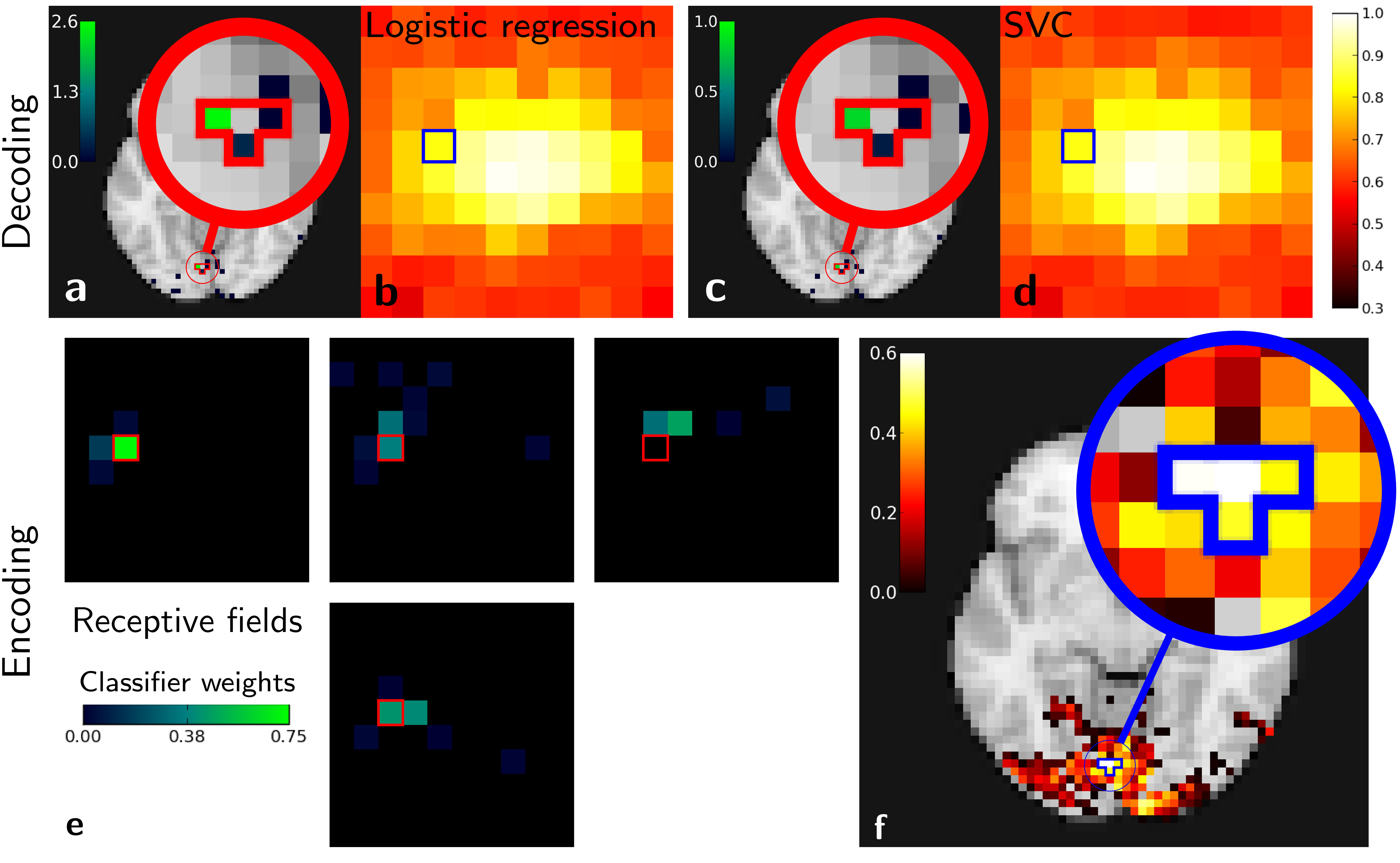}
  \end{center}
  \caption{
      Miyawaki results in both decoding and encoding. Relations between one
      pixel and four brain voxels is highlighted for both methods.
      \textbf{Top: Decoding.} Classifier weights for the pixel
      highlighted (\textit{a.} Logistic regression, \textit{c.} SVM).
      Reconstruction accuracy per pixel
      (\textit{b.} Logistic
      regression, \textit{c.} SVM). 
      \textbf{Bottom: Encoding.} \textit{e}: receptive fields corresponding to
       voxels with highest scores and its neighbors.
       \textit{f}: reconstruction accuracy depending on 
	  pixel position in the stimulus. --- Note that the pixels and voxels highlighted are the same in 
      both decoding and encoding figures and that encoding and decoding
      roughly match as both approach highlight a relationship between the
      same pixel and voxels.
}
\label{fig:miyawaki}
\end{figure}

Figure~\ref{fig:miyawaki} gives encoding and decoding results: the relationship
between a given image pixel and four voxels of interest in the brain.
In decoding settings, Figures \ref{fig:miyawaki}\textit{a} and \ref{fig:miyawaki}\textit{c} show the classifier's weights as brain
maps for both methods. They both give roughly the same results and we can
see that the weights are centered in the V1 and nearby retinotopic areas.
Figures \ref{fig:miyawaki}\textit{b} and \ref{fig:miyawaki}\textit{d} show reconstruction
accuracy score using Logistic Regression (LR) and SVM
(variable \texttt{mean\_scores} in the code above).
Both methods give almost identical results. 
As in the original work \citep{miyawaki2008}, reconstruction is more
accurate in the fovea.
This is explained by the higher density of neurons dedicated to
foveal representation in the primary visual area.

In encoding settings, figure~\ref{fig:miyawaki}\textit{e} shows classifiers
weights for encoding, that we interpret as receptive fields. We can
see that receptive fields of neighboring voxels are neighboring
pixels, which is expected from retinotopy: primary visual
cortex maps the visual field in a topologically organized manner.

Both encoding and decoding analysis show a link between the selected
pixel and brain voxels. In the absence of ground truth, seeing that
different methods come to the same conclusion comes as face validity.

\section{Resting-state and functional Connectivity analysis}

Even in the absence of external behavioral or clinical variable, studying
the structure of brain signals can reveal interesting information.
Indeed, \cite{biswal1995} have shown that brain activation exhibits
coherent spatial patterns during rest. These correlated voxel activations
form functional networks that are consistent with known task-related networks
\citep{smith2009}.

Biomarkers found via predictive modeling on resting-state fMRI would be
particularly useful, as they could be applied to diminished subjects that
cannot execute a specific task. Here we use a dataset containing control
and ADHD (Attention Disorder Hyperactivity Disorder) patients resting
state data (subjects are scanned without giving them any specific task to
capture the cerebral background activity).

Resting state fMRI is unlabeled data in the sense that the brain activity
at a given instant in time cannot be related to an output variable.
In machine learning, this class of problems is known as unsupervised
learning. 
To extract functional networks or regions, we use methods that group together 
similar voxels by comparing their time
series. In neuroimaging, the most popular method is ICA that
is the subject of our first example. We then show how to obtained 
functionally-homogeneous regions with
clustering methods.

\subsection{Independent Component Analysis (ICA) to extract networks}

ICA is a blind source separation method. Its principle is to separate a
multivariate signal into several components by maximizing their non-Gaussianity.
A typical example is the \emph{cocktail party problem} where ICA is able to separate
voices from several people using signal from microphones located across the room.

\subsubsection{ICA in neuroimaging}

ICA is the reference method to extract networks from resting state
fMRI \citep{kiviniemi2003}. Several strategies have been used to syndicate ICA
results across several subjects. \cite{calhoun2001a} propose a dimension
reduction (using PCA) followed by a concatenation of timeseries (used in this
example). \cite{varoquaux2010} use dimension reduction and canonical correlation analysis
to aggregate subject data. Melodic \citep{beckmann2004}, the ICA tool in
the FSL suite, uses a concatenation approach not detailed here.

\subsubsection{Application}

As data preparation steps, we not only center, but also detrend the time
series to avoid capturing linear trends with the ICA. Applying to the
resulting time series the FastICA algorithm~\citep{Hyvarinen:2000vk} with scikit-learn is
straightforward thanks to the transformer concept. The data matrix must
be transposed, as we are using \emph{spatial} ICA, in other words the
direction considered as random is that of the voxels and not the time
points. The maps obtained capture different components of the signal,
including noise components as well as resting-state functional networks.
To produce the figures, we extract only 10 components, as we are
interested here in exploring only the main signal structures.

\begin{lstlisting}
# Here we start with Xs: a list of subject-level data matrices
# First we concatenate them in the time-direction, thus implementing
# a concat-ICA
X = np.vstack(Xs)
from sklearn.decomposition import FastICA
ica = FastICA(n_components=10)
components_masked = ica.fit_transform(data_masked.T).T
\end{lstlisting}

\subsubsection{Results}

\begin{figure}[hbtp]
  \centerline{\includegraphics[width=.9\linewidth]{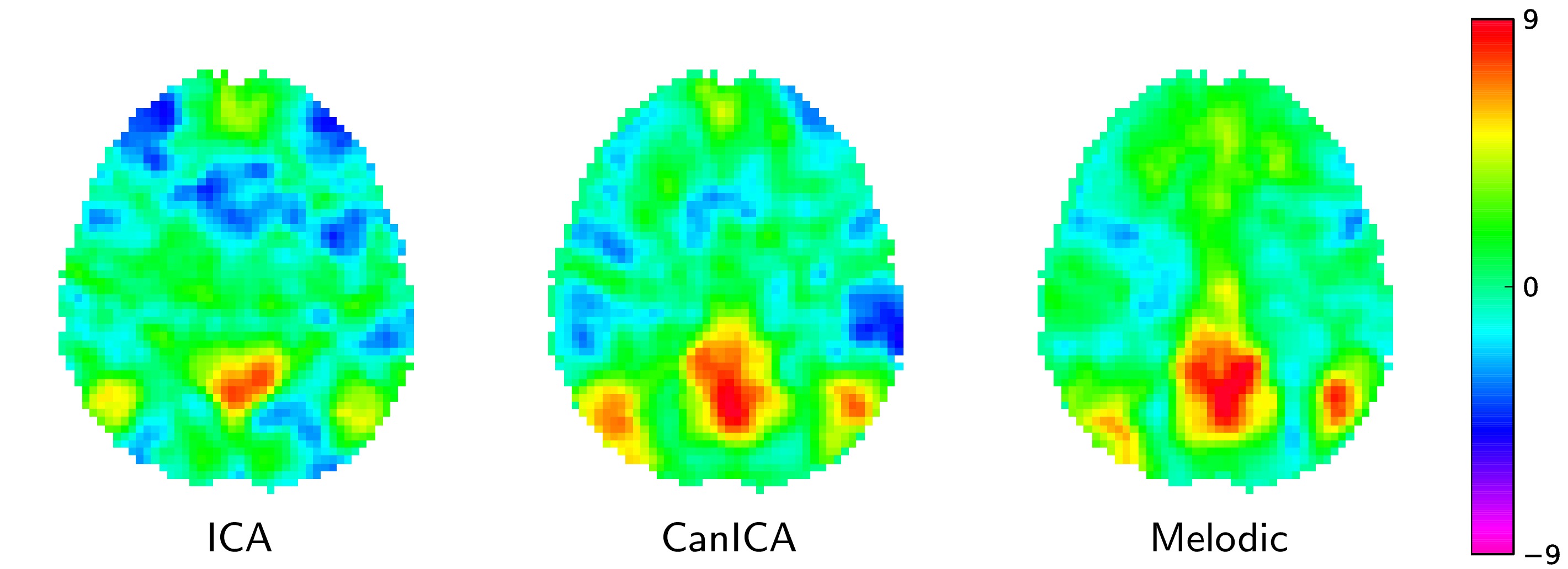}}
  \caption{Default mode network extracted using different approaches:
\emph{left}: the simple Concat-ICA approach detailed in this article;
\emph{middle}: CanICA, as implemented in nilearn; \emph{right}: Melodic's
concat-ICA. Data have been normalized (set to unit variance) for display
purposes.}
  \label{fig:ica}
\end{figure}

On fig. \ref{fig:ica} we compare a simple concat ICA as implemented by
the code above to more sophisticated multi-subject methods, namely Melodic's
concat ICA and CanICA--also implemented using scikit-learn although we do
not discuss the code here. We display here only the default mode network
as it is a well-known resting-state network. It is hard to draw conclusions from
a single map but, at first sight, it seems that both CanICA and Melodic
approaches are less subject to noise and give similar results.

Scikit-learn proposes several other matrix decomposition strategies listed in
the module `sklearn.decomposition`. A good alternative to ICA is the dictionary
learning that applies a $\ell_1$ regularization on the extracted components
\citep{varoquaux2011}.
This leads to more sparse and compact components than ICA ones, which are
full-brain and require thresholding.

\subsection{Learning functionally homogeneous regions with clustering}
\label{clustering}

From a machine learning perspective, a clustering method aggregates 
samples into groups (called clusters) maximizing a measure of similarity
between samples within each cluster. If we consider voxels of a functional brain image
as samples, this 
measure can be based on functional similarity, leading to clusters of voxels
that form functionally homogeneous regions \citep{thirion2006}.

\subsubsection{Approaches}

Several clustering approaches exists, each one having its own pros and
cons. Most require setting the number of clusters extracted. This choice
depends on the application: a large number of clusters will give a more
fine-grained description of the data, with a higher fidelity to the
original signal, but also a higher model complexity. Some clustering
approaches can make use of spatial information and yield
spatially contiguous clusters, \emph{i.e.} parcels. Here we will describe
two clustering approaches that are simple and fast.

\paragraph{Ward clustering} uses a bottom-up hierarchical approach:
voxels are progressively agglomerated together into clusters. In
scikit-learn, structural information can be specified via a connectivity
graph given to the Ward clustering estimator. This graph is used to allow
only merges between neighboring voxels, thus readily producing contiguous
parcels. We will rely on the {\tt
sklearn.feature\_extraction.image.grid\_to\_graph} function to
construct such a graph using the neighbor structure of an image grid,
with optionally a brain mask.

\paragraph{K-Means} is a more top-down approach, seeking cluster centers
to evenly explain the variance of the data. Each voxels are then assigned
to the nearest center, thus forming clusters. As imposing a spatial model
in K-means is not easy, it is often advisable to spatially smooth the
data.

To apply the clustering algorithms, we run the common data preparation
steps and produce a data matrix. As both Ward clustering and K-means rely
on second-order statistics, we can speed up the algorithms by reducing
the dimensionality while preserving these second-order statistics with a
PCA. Note that clustering algorithms group samples and that here we want
to group voxels. So if the data matrix is, as previously a (time points
$\times$ voxels) matrix, we need to transpose it before running the
scikit-learn clustering estimators. Scikit-learn provides a
\texttt{WardAgglomeration} object to do this \emph{feature agglomeration}
with Ward clustering \citep{michel2012supervisedclustering}, but this is
not the case when using K-Means.

\begin{lstlisting}
connectivity = grid_to_graph(n_x=mask.shape[0], n_y=mask.shape[1],
                             n_z=mask.shape[2], mask=mask)
ward = WardAgglomeration(n_clusters=1000, connectivity=connectivity)
ward.fit(X)
# The maps of cluster assignment can be retrieved and unmasked
cluster_labels = numpy.zeros(mask.shape, dtype=int)
cluster_labels[mask] = ward.labels_
\end{lstlisting}

\subsubsection{Results}

\begin{figure}[hbtp]
  \includegraphics[width=\linewidth]{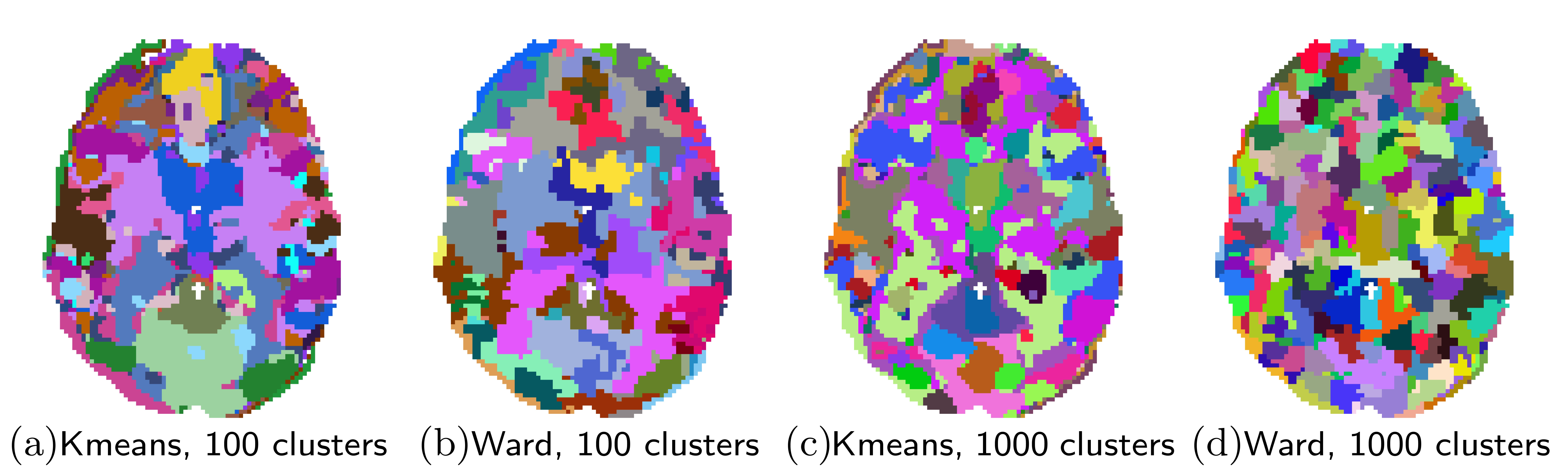}
  \caption{Brain parcellations extracted by clustering. Colors are
random.}
  \label{fig:clustering}
\end{figure}

Clustering results are shown in figure~\ref{fig:clustering}. While
clustering extracts some known large scale structure, such as the
calcarine sulcus on fig~\ref{fig:clustering}.a, it is not guaranteed to
delineate functionally specific brain regions. Rather, it can be considered as a compression, that
is a useful method of summarizing information, as it groups together
similar voxels. Note that, as K-means does not extract spatially-contiguous
clusters, it gives a number of regions that can be much larger than the
number of clusters specified, although some of these regions can be very
small. On the opposite, spatially-constrained Ward directly creates regions.
As it is a bottom-up process, it tends to perform best
with a large number of clusters. There exist many more clustering
techniques exposed in scikit-learn. Determining which is the best one to
process fMRI time-series requires a more precise definition of the target
application.

Ward's clustering and K-Means are among the simplest approaches proposed in the
scikit-learn. \cite{craddock2011} applied spectral clustering on neuroimaging
data, a similar application is available in nilearn as an example.

\section{Conclusion}

In this paper we have illustrated with simple examples how machine
learning techniques can be applied
to fMRI data using the scikit-learn Python toolkit in order to tackle
neuroscientific problems. Encoding and decoding can rely on supervised
learning to link brain images with stimuli. Unsupervised learning
can extract structure such as functional networks or
brain regions from resting-state data. The accompanying Python code for the machine learning
tasks is straightforward. Difficulties lie in applying proper 
preprocessing to the data, choosing the right model for the problem,
and interpreting the results. Tackling these difficulties while
providing the scientists with simple and readable code requires building
a domain-specific library, dedicated to applying scikit-learn to
neuroimaging data. This effort is underway in a nascent project, nilearn,
that aims to facilitate the use of scikit-learn on neuroimaging data.

The examples covered in this paper only scratch the
surface of applications of statistical learning to neuroimaging.
The tool stack presented here shines uniquely in this regard as it opens the
door to any combination of the wide range of machine learning methods
present in scikit-learn with neuroimaging-related code. For
instance, sparse inverse covariance can extract the functional 
interaction structure from fMRI time-series \citep{varoquaux2013} using
the graph-lasso estimator.
Modern neuroimaging data analysis entails fitting rich models on
limited data quantities. These are high-dimensional statistics problems
which call
for statistical-learning techniques. We hope that bridging a
general-purpose machine learning tool, scikit-learn, to domain-specific
data preparation code will foster new scientific advances.

\section*{Disclosure/Conflict-of-Interest Statement}
The authors declare that the research was conducted in the absence of any
commercial or financial relationships that could be construed as a potential
conflict of interest.

\paragraph{Funding\textcolon} We acknowledge funding from the NiConnect
project and NIDA R21 DA034954, SUBSample project from the DIGITEO
Institute, France.

\bibliographystyle{frontiersinSCNS} 
\bibliography{biblio}

\end{document}